# *OntoSeg*: a Novel Approach to Text Segmentation using Ontological Similarity

Mostafa Bayomi, Killian Levacher, M.Rami Ghorab, Séamus Lawless,
Centre for Global Intelligent Content, Knowledge and Data Engineering Group
School of Computer Science and Statistics, Trinity College Dublin,
Dublin, Ireland
{bayomim, killian.levacher, rami.ghorab, seamus.lawless}@scss.tcd.ie

*Abstract*— Text segmentation (TS) aims at dividing long text into coherent segments which reflect the subtopic structure of the text. It is beneficial to many natural language processing tasks, such as Information Retrieval (IR) and document summarisation. Current approaches to text segmentation are similar in that they all use word-frequency metrics to measure the similarity between two regions of text, so that a document is segmented based on the lexical cohesion between its words. Various NLP tasks are now moving towards the semantic web and ontologies, such as ontology-based IR systems, to capture the conceptualizations associated with user needs and contents. Text segmentation based on lexical cohesion between words is hence not sufficient anymore for such tasks. This paper proposes *OntoSeg*, a novel approach to text segmentation based on the ontological similarity between text blocks. The proposed method uses ontological similarity to explore conceptual relations between text segments and a Hierarchical Agglomerative Clustering (HAC) algorithm to represent the text as a tree-like hierarchy that is conceptually structured. The rich structure of the created tree further allows the segmentation of text in a linear fashion at various levels of granularity. The proposed method was evaluated on a well-known dataset, and the results show that using ontological similarity in text segmentation is very promising. Also we enhance the proposed method by combining ontological similarity with lexical similarity and the results show an enhancement of the segmentation quality.

## I. INTRODUCTION

Text segmentation is the process of placing boundaries within text to create segments according to some task-dependent criterion. An example of text segmentation is topical segmentation, which aims to segment a text according to the subjective definition of what constitutes a topic. Text segmentation algorithms are widely used as an essential step in many Natural Language Processing (NLP) tasks, such as Information Retrieval [1] [2], document summarisation [3], Question answering [4] and Automatic generation of E-Learning Courses [5]. In Information Retrieval, a document is segmented into distinct topics and only the topical segments relevant to the user's needs are retrieved. Segmentation not only provides more accurate information to the user, but also reduces the user's burden to read the whole document. In document summarisation, a document is segmented into topics and then each topic is summarized independently. This process guarantees that the final summary covers all the key topics in the document.

There are different approaches to text segmentation in the literature. Some approaches segment text linearly [6] and others segment it hierarchically[7]. *TextTiling*, for example, is a well-known linear text segmentation method proposed by Hearst [8] that measures the lexical similarity between text blocks. Text blocks are the smallest units that constitute the text. They range from one sentence [9] to multiple sentences (paragraphs) [10]. *TextTiling* uses a sliding window and follows the peaks and valleys of the similarity curve to determine where to segment a text. Utiyama and Isahara [11] segmented all possible partitions using dynamic programming and used the probability distribution of the words to rank and select the best segments.

These methods are similar in that they all use word frequency metrics to measure the similarity between two regions of text so that a document is segmented at the points where the connections between the regions of words are the weakest, which means that the obtained segments from these approaches are segmented based on the lexical relationship between words in the text. As mentioned before, text segmentation is an essential step for many NLP tasks; these NLP tasks are moving now towards the Semantic Web and the use of ontologies. In Information Retrieval, for example, systems that are based on keywords provide limited capabilities to capture the conceptualizations associated with user needs and contents. In order to solve these limitations, the idea of semantic searches, based on the conceptual meaning of text, has been the focus of a wide body of research and many ontology-based IR systems have been developed [12]. In such systems, whereby text is segmented based solely on the relation between words, such method represents a limitation to capture the conceptualizations associated with user needs. Hence, a need for segmenting and representing text based on the ontological relation between its constituents arises.

In this paper, we propose *OntoSeg*, a novel approach to text segmentation based on the ontological similarity between text blocks. In contrast to traditional text segmentation approaches which used lexical-based similarity between words, we use ontology-based similarity to assess the relatedness between text blocks. A Hierarchical Agglomerative Clustering (HAC) approach is then applied to cluster similar blocks together. The output is a hierarchy that is constructed based on how text blocks (one or more sentence) are conceptually related to each other. Our experiments demonstrate that segmenting text based on the ontological similarity is applicable with a low error rate.

This research has three contributions:
1- Segmenting text based on the ontological similarity between text blocks (as opposed to lexical similarity); this method is intended for use in ontology-based NLP tasks.
2- A method aimed at enhancing the quality of segments produced when no ontological relation between text blocks exists is also presented
3- Evaluating the quality of text segmentation using the ontology-based similarity.

The rest of the paper is organized as follows. In Section II, recent related work in the text segmentation literature is reviewed. Section III presents the proposed ontological text segmentation model. Section IV describes the experiments and the dataset that were used to evaluate the proposed method. Section V describes the evaluation process and the evaluation metrics used therein. Results and findings are discussed in Section VI. Section VII concludes the paper with some future research directions.

## II. RELATED WORK

Various synonyms in the literature are used to refer to text segmentation such as: Linear Text Segmentation [6], Hierarchical Text Segmentation [7], Topic Segmentation [13][14], Text Boundaries or Boundary Determination [15], and Topic Boundaries [16].

Furthermore, segmentation tasks have been categorised from different points of view as:
1) Content-based and Discourse-based.
2) Supervised and Unsupervised.
3) Linear and Hierarchical.
4) Borderline sentences detection methods:
   a) Similarity based methods.
   b) Graphical methods.
   c) Lexical chain based methods.

### A. Content-based and Discourse-based

Content-based approaches focus on the story content and resolve the segmentation problem by relying on some measure of the difference in word usage on the two sides of a potential boundary: the larger the difference, the more indicative of a boundary. A well-known content-based approach example is *TextTiling* proposed by Hearst [8]. TextTiling is content-based text segmentation algorithm that uses a sliding window approach to segment a text. The calculation is accomplished by two vectors containing the number of occurring terms of each block. The similarities between adjacent blocks within the text are computed to detect topic changes. The computed similarities are smoothed, and used to identify topic boundaries by a cutoff function.

On the other hand, the discourse-based techniques focus on story structure or discourse. These approaches make use of prosodic features such as pause duration as well as lexical features such as the presence of certain cue phrases that tend to appear near the segment boundaries. An example of discourse-based approaches is the Hidden Markov Model (HMM) segmentation method [17] that models "marker words", or words which predict a topic change.

### B. Supervised and Unsupervised.

A supervised text segmentation approach called *divSeg* was introduced by Song et al. [18], where they apply an iterative approach that splits text at its weakest point in terms of the lexical connectivity strength between two adjacent parts. After they found the weakest point in the text, their approach produces a deep and narrow binary tree. The tree is then flattened into a broad and shallow hierarchy through supervised learning of a document set or explicit input of how a text should be segmented. Hsueh et al. [19] described a supervised hierarchical topic segmentation approach that trains separate classifiers for topic and sub-topic segmentation.

On the other hand, Eisenstein and Barzilay[14] proposed a Bayesian approach to unsupervised topic segmentation. They showed that lexical cohesion between text segments can be placed in a Bayesian context by modelling the words in each topic segment. *TextTiling* [8] and *C99* [20] are also considered unsupervised linear topic segmentation algorithms.

### C. Linear and Hierarchical

If we look at the text segmentation from the text representation perspective, we can divide it into linear and hierarchical approaches. Linear text segmentation deals with the sequential analysis of topical changes where segments are non-overlapping and sequential. It has been argued that this sequence model is sufficient for many purposes [8]. An early linear text segmentation algorithm was the *TextTiling* approach introduced by Hearst [10] in 1997. Galley et al. [21] proposed *LcSeg*, a *TextTiling*-based algorithm that uses *tf-idf* term weights, which improves text segmentation results. Another well-known linear text segmentation algorithm is *C99* introduced by Choi [20]. *C99* segments a text by combining a rank matrix, transformed from the sentence-similarity matrix, and divisive clustering. Choi et al. [22] introduced another enhanced version of *C99* by applying Latent concept modelling to the similarity metric. They showed that using a Latent Semantic Analysis (LSA) based metric could improve the segmentation accuracy. Utiyama and Isahara [11] introduced probabilistic approaches using Dynamic Programming (DP) called *U00*. DP can be used to efficiently find paths of minimum cost in a graph. DP is used in text segmentation to represent each possible segment (e.g. every sentence boundary) as an edge providing a cost function that penalizes common vocabulary across segment boundaries. Misra et al. [23] used Latent Dirichlet Allocation (LDA) topic model to linearly segment a text into semantically coherent segments.

Hierarchical text segmentation concerns itself with finding more fine grained subtopic structures in texts. The first hierarchical algorithm was proposed by Yaari [7]. Yaari used paragraphs as an elementary units for his algorithm and he measured the cohesion between them using cosine similarity. An agglomerative clustering approach is then applied to induce a dendrogram over paragraphs; the dendrogram is subsequently transformed into a hierarchical segmentation. A hierarchical Bayesian algorithm based on LDA is introduced by Eisenstein [24].

*D. Borderline sentences detection methods*

There are three main approaches to detect borderline sentences within text [2]:

*1) Similarity based methods:* represent text blocks as vectors and then measure the proximity by using (most of the time) the cosine of the angle between these vectors. The *C99* algorithm [20] for example uses a similarity matrix to generate a local classification of sentences and isolate topical segments.

*2) Graphical methods:* represent terms frequencies and use these representations to identify topical segments (which are dense dot clouds on the graphic). The *DotPlotting* algorithm [16] is the most common example of the use of a graphical approach of text segmentation.

*3) Lexical chains based methods:* the notion of lexical chains was first proposed by Morris and Hirst [25] to chain semantically related words together via a thesaurus. A chain links multiple occurrences of a term in the document and is considered broken when there are too much sentences between two occurrences of a term. *Segmenter* [26] uses this method for text segmentation with a subtle adjustment as it determines the number of necessary sentences to break a chain in function of the syntactical category of the term.

All the aforementioned approaches have focused on the similarity (or dissimilarity) between text blocks based on the words that constitute the text. Even the approaches that relied on semantic analysis only applied a shallow semantic parsing of text to discover different kinds of relationships between two words, including synonymy (the same meaning) and hyponymy (where one word is a more specific instance of another). Our research therefore focuses on semantically mining text and applying deep semantic analysis of text to discover the relation between its constituents.

In our approach we measure the similarity between text blocks based on the ontology-based semantic similarity. The ontology-based semantic similarity relates to computing the similarity between conceptually similar but not necessarily lexically similar terms. Semantic similarity has been widely used in many research fields such as: (1) *Information Retrieval*: to improve accuracy of current Information Retrieval techniques and semantic indexing [12]. (2) *Natural Language Processing tasks*: there are several tasks such as word sense disambiguation [27], synonym detection [28], sentiment analysis [29], analogical reasoning for sentiment analysis [30] or automatic spelling error detection and correction [31]. (3) *Knowledge management*: such as thesauri generation [32], information extraction [33], semantic annotation [34] and ontology merging and learning [35], in which new concepts should be discovered or acquired from text in order to relate them to already existing ones.

Ontology-based similarity can be classified into three main approaches:

*1) Edge-counting approaches:* where a straightforward method to calculate similarity between two concepts is to compute the minimum path length connecting their corresponding ontological nodes via is-a links [36]. The longer the path, the more semantically far the terms are.

*2) Feature-based approaches:* on the contrary to edge-counting approaches, feature-based approaches assess similarity between concepts as a function of their properties [28]. They take into account common and noncommon features of the compared terms.

*3) Information Content (IC) based approaches:* these approaches are associated with the appearance probabilities of each concept in the taxonomy computed from their occurrences in a given corpus. IC of a term is computed according to the negative log of its probability of occurrence. In this manner, infrequent words are considered more informative than common ones.

In this research, we rely on an Edge-counting approach proposed by Wu and Palmer [36] as its performance is deemed better than other methods [28].

III. SEGMENTATION BY HIERARCHICAL AGGLOMERATIVE CLUSTERING

The segmentation process proposed in this paper consists of three phases:

1) Semantic annotation.

2) Calculating similarity between text blocks (sentences or paragraphs).

3) Hierarchical Agglomerative Clustering (HAC).

*A. Semantic annotation:*

In this phase, the text is semantically annotated using a named entity recognition algorithm and text entities are extracted. Each entity is then mapped to its class or classes in an ontology and the text is represented as a sentence-based vector-space. This vector space is then used as an input to the following phase. Several ontologies exist nowadays, some of them are domain-specific ontologies (such as the *MeSH[1]* ontology of medical and biomedical terms), while others are cross-domain (such as *DBpedia[2]*). As we are not focusing on a specific domain, in this research we use DBpedia ontology as the underlying knowledge base, as opposed to a domain-specific alternative. *DBpedia Spotlight[3]* is used as the named entity recognition system to extract entities from the targeted text. DBpedia Spotlight is a tool for automatically annotating mentions of DBpedia resources in text, providing a solution for linking unstructured information sources to the Linked Open Data cloud through DBpedia. DBpedia Spotlight recognizes entities that have been mentioned in text and subsequently matches these entities to their classes in DBpedia ontology. For each annotated entity in the text, the classes that match this entity are extracted. For example, *Barack Obama*, as an entity, matches with DBpedia classes: *["Person", "Agent", "Officeholder"]*. Since the elementary blocks for the proposed approach are sentences, each sentence in the text is represented as a vector of entities, and each entity is represented by a set of classes that match the entity from DBpedia. A sentence based vector space is built and a similarity between its adjacent vectors is measured as discussed in the following subsection.

---

[1] http://www.nlm.nih.gov/mesh
[2] http://dbpedia.org/
[3] https://github.com/dbpedia-spotlight/dbpedia-spotlight/wiki

## B. Similarity Computation:

The key idea proposed in this research consists in treating the segmentation of text based on the ontological similarity between its blocks. A text block is the elementary unit to the segmentation algorithm, which could be one sentence or multiple sentences (paragraphs).

We measure the similarity between text units based on two similarity measures: (1) Ontological similarity and (2) Lexical similarity.

### 1) Ontological similarity:

To measure the ontological similarity between two text blocks, we measure the similarity between the classes of their entities using the **is-a** relation. In ontology structure, the **is-a** relations group the classes according to how they are conceptually related to each other. Given a pair of two classes, $c_1$ and $c_2$, a well-known method with intuitive explicitness for assessing their similarity is to calculate the distance between these classes in an ontology hierarchy; the shorter the distance, the higher the similarity. In the case that multiple paths between the nodes exist, the shortest distance of all paths is used.

Several measures have been developed for measuring similarity between two concepts in a taxonomy. Out of these, we choose the measure proposed by Wu and Palmer [36] because it has shown performance improvements over other methods [28]. The principle behind Wu and Palmer's similarity computation is based on the edge-counting method, whereby the similarity of two concepts is defined by how closely they are related in the hierarchy, i.e., their structural relations. Given two concepts $c_1$ and $c_2$, the conceptual similarity between them is:

$$ConSim(c_1, c_2) = 2*N/(N_1+N_2) \quad (1)$$

where N is the distance between the closest common ancestor (CS) of $c_1$ and $c_2$ and the taxonomy root, and $N_1$ and $N_2$ are the distances between the taxonomy root on one hand and $c_1$ and $c_2$ on the other hand respectively.

The similarity between two entities can be defined as a summation of weighted similarities between pairs of classes in each of the entities. Given two entities $E_1$ and $E_2$, the similarity between them is:

$$EntSim(E_1, E_2) = \frac{\sum_{i=1}^{m}\sum_{j=1}^{n} ConSim(c_i, c_j)}{m \times n} \quad (2)$$

where *m* and *n* are the two sets of classes that *E1* and *E2* have respectively.

Equation (2) calculates the similarity between two entities, where each entity belongs to one or more classes. For example, *Barack Obama* as an entity is mapped to three DBpedia classes: *["Person", "Agent", "Officeholder"]*, and *George Bush* also is mapped to three DBpedia classes: *["Person", "Agent", and "Officeholder"]*. Hence, although the two entities are not lexically similar, or even close to each other, they are deemed ontologically similar. This is the idea behind the ontological similarity: it measures the similarity between entities according to the conceptual characteristics which they share. As another example of how ontological similarity differentiates between entities, consider *Michael Jackson* as an entity that is mapped to four DBpedia classes: *["Person", "Agent", "Artist", "MusicalArtist"]*. Intuitively, the two entities *Barack Obama* and *George Bush* are more ontologically similar to each other than either of them is to *Michael Jackson*.

On a text-block level, the similarity between two blocks can be defined as a summation of weighted similarities between pairs of entities in each of the units.

Given two text blocks *B1* and *B2*, which have a set of entities *a* and *b* respectively, the similarity between *B1* and *B2* is:

$$BlockSim(B1, B2) = \frac{\sum_{i=1}^{a}\sum_{j=1}^{b} EntSim(E_i, E_j)}{a \times b} \quad (3)$$

### 2) Lexical similarity:

Lexical similarity has been used widely in the literature in text segmentation [8], [20], and as its name suggests, it splits text into segments that are lexically coherent. Lexical cohesion refers to the connectivity between two portions of text in terms of word relationships.

Although text blocks might share ontological similarities between each other, it may be the case that ontological similarity alone is not sufficient to measure how text blocks are coherent with each other. This is due to the following reasons:

1- Text blocks might not contain any entities at all.

2- The entity extraction algorithm may not discover some entities in the text block.

3- The extracted entities from a text block may not be sufficient to reflect the similarity between text blocks.

4- The used ontology may not cover all the text mentions.

Thus, the lexical overlap between text blocks should be part of the overall similarity measure. As a result, we enrich our similarity measure by obtaining the lexical similarity between text blocks and combine it with the ontological similarity. To measure the lexical similarity between text blocks, first, stopwords are removed from the text as they are generally assumed to be of less, or no, informational value. Then the remaining words are stemmed and each block is represented by a lexical frequency vector. A lexical vector cosine similarity is calculated. It is defined as the cosine of the angle between two vectors $v$ and $w$ such that:

$$cos\,\theta = \frac{\sum_{i=1}^{n} v_i\, w_i}{\sqrt{\sum_{i=1}^{n}(v_i)^2} \cdot \sqrt{\sum_{i=1}^{n}(w_i)^2}} \quad (4)$$

### C. Hierarchical Agglomerative Clustering (HAC)

Hierarchical clustering algorithms have been studied extensively in the clustering literature [37]. The general concept of agglomerative clustering is to successively merge documents into clusters based on their similarity with one another. The agglomerative clustering technique could be transferred from document level into text level, where the clustering process is done between text blocks, within a document (as opposed to across whole documents) [7]. When applying Hierarchical Agglomerative Clustering on text blocks the algorithm successively agglomerates blocks that are coherent to each other, thus forming a text structure.

The idea behind using HAC in text segmentation is that it is a bottom-up clustering approach, which means that it starts from the smallest chunks and then builds the text hierarchy by merging text blocks (clusters) based on how near or similar they are to each other. In contrast, the top-down (divisive) clustering approach starts from the full document and then divides the text into smaller blocks based on how far (i.e. how different) they are from each other. Hence, the output of the bottom-up approach can be regarded as hierarchically coherent tree. Thus, the method of Hierarchical Agglomerative Clustering for text is useful to support a variety of search methods because it naturally converts text into a tree-like hierarchy and provides different levels of granularity for the underlying content; this can then easily be leveraged for the search process.

Unlike general HAC for clustering documents, where at each stage the proximity of the newly merged object to all other available segments is computed, on text level we compute only the similarity of the text block to its two neighbours. This is because we require that the linear order in the text be preserved in the structure. The implication on complexity is that while general HAC algorithm for documents takes an order of $O(N^2)$ steps, it takes only $O(N)$ on text level.

The algorithm successively clusters "coherent" segments based on the accumulation between the ontological and lexical similarity scores between text blocks, which guarantees the ontological and lexical cohesion between agglomerated segments. The HAC algorithm for text segmentation, based on blocks as the elementary segments, is shown in *Fig. 1*.

**Input:** *Ontological and Lexical space models*
**Output:** *the partitioned segments as a tree*
**begin**
   **while** (number of blocks>1) **do**
      **foreach** *block Bi* **do**
         *calculate similarity between $B_i$ and its neighbors*
         Ontological Similarity: Osim ($B_i$, $B_{i+1}$) and Osim ($B_i$, $B_{i-1}$)
         Lexical Similarity: Lsim ($B_i$, $B_{i+1}$) and Lsim ($B_i$, $B_{i-1}$)
         **Total Similarity:**
         Tsim ($B_i$, $B_{i+1}$) = Osim ($B_i$, $B_{i+1}$) + Lsim ($B_i$, $B_{i+1}$)
         Tsim ($B_i$, $B_{i-1}$) = Osim ($B_i$, $B_{i-1}$) + Lsim ($B_i$, $B_{i-1}$)
        **if** (sim($B_i$, $B_{i+1}$) >= sim($B_i$, $B_{i-1}$)) **then**
            merge ($B_i$, $B_{i+1}$)
        **end**
        **else**
            merge ($B_i$, $B_{i-1}$)
        **end**
      **end**
   **end**
**end**

Fig. 1. Hierarchical Agglomerative Clustering of text segments

Conceptually, the process of agglomerating blocks into successively higher levels of clusters creates a cluster hierarchy (dendrogram) for which the leaf nodes correspond to individual blocks, and the internal nodes correspond to the merged groups of clusters. When two groups are merged, a new node is created in this tree corresponding to this larger merged group. The two children of this node correspond to the two groups of blocks which have been merged to it. *Fig. 2* shows the resulted dendrogram from the algorithm for a sample text.

### D. From hierarchical into linear representation

The hierarchical text segmentation produces a tree that can be used as a visual illustration of the underlying hierarchical structure of a document. *Fig. 3* depicts a tree representation of a sample text of 10 sentences. The benefit of this tree is that it represents different levels of granularity of the document, which in turn means that the document can be segmented into different segmentation levels. This is a powerful criterion in the hierarchical representation of text. In contrast to linear representation, in each level of the structure (tree) segmentation with different levels of details could be obtained and can be usefully applied to many other tasks' needs.

In order to convert a hierarchical representation into a linear representation a threshold corresponding to the number of the segments needed is set and the level that contains the corresponding number of nodes in the tree is extracted. If this number is not represented in one of the tree levels, a flattening process is applied to the largest nodes. For example, suppose that the specified number is 10 segments, and in one of the tree levels the number of nodes (segments) is seven nodes. As now we need three more segments, for the largest three nodes (large in number of blocks) they are flattened by obtaining the two subsequent nodes that constitute this large node, i.e. we go down a level in the tree for those three large segments. This method of flattening the tree guarantees that the coherency between the obtained segments is preserved.

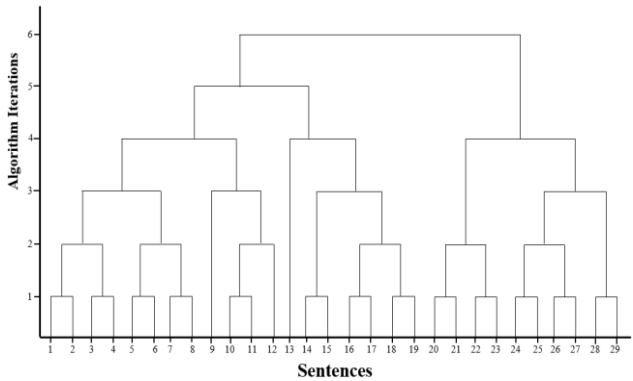

Fig. 2. Sentences dendrogram of a sample text.

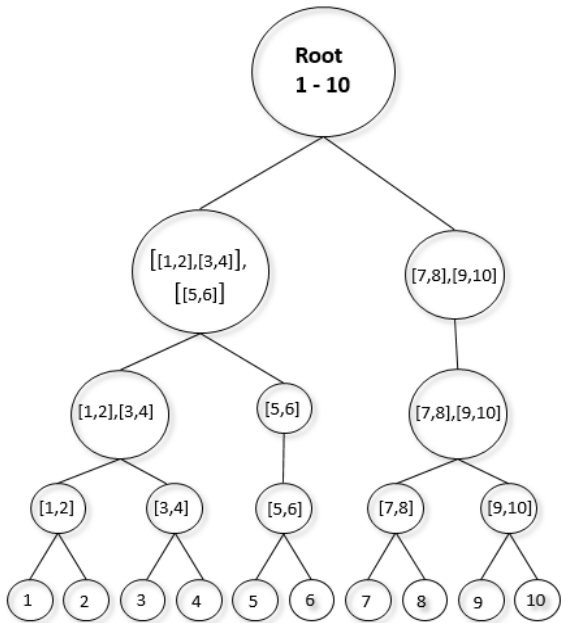

Fig. 3 A tree representation for a text from 10 sentences

## IV. EXPERIMENTAL SETUP

The output from the proposed approach is a tree that represents the text hierarchy. As depicted in *Fig 2.*, each level in the tree represents a level of granularity for the text where each node, in that level, represents a segment that contains coherent blocks. As mentioned before, a linear representation of text can be obtained from such a tree, which means that our method can be evaluated as a linear text segmentation method. In this experiment, we evaluated the efficiency of our approach on Choi's dataset[4] [20]. This dataset has been widely used in linear text segmentation evaluation [38][39]. The dataset consists of documents made up of ten concatenated text segments. Each segment consists of the first *n* sentences of a randomly selected document from the Brown Corpus. The dataset is divided into four subsets and are listed in the table below. There are a total of 700 text documents.

---

[4] Choi's C99 release and the dataset are available here :
http://web.archive.org/web/20040810103924/http://www.cs.man.ac.uk/~mary/choif/software.html

TABLE I.    TEST DATASET STATISTICS

| Range of *n* | 3-11 | 3-5 | 6-8 | 9-11 |
|---|---|---|---|---|
| # samples | 400 | 100 | 100 | 100 |

Each document in the dataset is processed and two vector space models are generated: the ontological and the lexical. Since the elementary text blocks to our method consists of sentences, each sentence in the ontological vector space is represented as a vector of sets of DBpedia classes where each set represents an entity that is extracted from the sentence. These sets of classes are used to measure the ontological similarity between sentence vectors according to (1), (2), and (3). To build the lexical vector space, first the stopwords are removed from the text and then the remaining terms are stemmed; after this, each sentence is subsequently represented as a term-frequency vector. The lexical similarity between adjacent vectors is then determined by calculating the cosine similarity between them (4).

A HAC algorithm is then applied on the obtained vector space models. For the ontological vector space, an ontological similarity score is calculated between each vector and its two neighbours. A lexical similarity score is also obtained for the lexical vector space. The final similarity score between two adjacent sentences is the combination of their ontological similarity and lexical similarity scores. For each set of three neighbouring sentences, the middle sentence is merged with the one that is most similar to it from the other two (e.g. sentence ***B*** is merged with ***C*** if the similarity score between ***B*** and ***C*** is higher than the score between ***A*** and ***B***). When the two neighbours are merged together they form a new text block (cluster) and two new vectors (ontological and lexical) are defined based on the new block to be used in the next iteration of the algorithm. Iteratively, the algorithm applies the same process between adjacent blocks until it merges all text blocks in one single cluster and a tree representation of the text is produced. A linear segmentation is then produced as mentioned before (Section III D) where the threshold is set to 10 as each document in Choi's dataset is consisting of 10 segments.

Since the main contribution of this research is to segment text based on the ontological similarity between its blocks, we first evaluate the quality of the produced segments based on the ontological similarity only. After that, we examine the impact of adding the lexical similarity to the ontological similarity using different weights for the two similarity measures.

The size of the elementary text blocks is considered a critical step in the segmentation process. Yaari [7] used paragraphs as the elementary blocks for his segmentation algorithm and affirms that the size of a paragraph, as opposed to a sentence, contains sufficient lexical information for the proximity test. Also Hearst et al. [8] measured the cosine similarity between text blocks where text blocks are consisting of fixed number of sentences (window). As a result, we examine the quality of the produced segments, using the ontological similarity only or the combination between the ontological and the lexical similarity, using varying window sizes: from one to four sentences.

According to the aforementioned considerations, we conducted four experimental runs (in each run, we used varying window sizes (1 to 4)):

*1) Experiment 1:* in the first run we use the ontological similarity only.

*2) Experiment 2:* in the second run, we use the combination between the ontological and lexical similarity scores with α = 0.3, where α specifies the weight of each of the two similarity measures. Let *Osim* and *Lsim* be the ontological and the lexical similarity scores respectively; the final score between two sentences (or blocks) S1 and S2 is:

$$Sim(S1, S2) = α * Lsim + (1- α) * Osim \quad (5)$$

Hence, α = 0.3 means that the ontological similarity score weight is 0.7 and the lexical score weight is 0.3.

*3) Experiment 3:* in this run treat both similarity scores equally, i.e. α = 0.5.

*4) Experiment 4:* in this run we give a higher weight to the lexical similarity by setting α=0.3.

## V. EVALUATION

It is common to evaluate text segmentation systems by *Pk* and / or *WindowDiff* measures. *Pk* and *WindowDiff* are penalty measurement metrics, which means that lower scores indicate higher segmentation accuracy. *Pk* was proposed by Beeferman et al. [40] as a measure that expresses a probability of segmentation error. To calculate *Pk*, we take a window of fixed width *k*, which is usually set to half of the average segment length in the reference partition, and move it across the segmented text, at each step examining whether the hypothesized segmentation is correct about the separation (or not) of the two ends of the window. *Pk* metric is defined as:

$$P_k = \sum_{1 \leq i \leq j \leq k} D_k(i,j) \left( \delta_{ref}(i,j) \oplus \delta_{hyp}(i,j) \right) \quad (6)$$

where δ*ref* (*i, j*) is an indicator function whose value is one if sentences *i* and *j* belong to the same segment and zero otherwise. Similarly, δ*hyp* (*i, j*) is one if the two sentences are hypothesized as belonging to the same segment and zero otherwise. The ⊕ operator is the XOR operator. The function *Dk* is the distance probability distribution that uniformly concentrates all its mass on the sentences which have a distance of *k*.

*WindowDiff* [41] is stricter as it not only decides whether there is a mismatch between the hypothesized partition and the reference partition, it also counts the difference of the number of segment boundaries in the given window between the two partitions. Thus, the results of *WindowDiff* are generally higher than those of *Pk* metric. *WindowDiff* is defined as:

$$WindowDiff(ref, hyp) = \frac{1}{K-k} \sum_{i=1}^{K-k} (|b(ref_i, ref_{i+k}) - b(hyp_i, hyp_{i+k})| > 0) \quad (7)$$

where *ref* is the correct segmentation for reference, *hyp* is the segmentation produced by the model, *K* is the number of sentences in the text, *k* is the size of the sliding window and *b(i, j)* is the number of boundaries between sentences *i* and *j*.

## VI. RESULTS

We evaluated our approach using the *WindowDiff* error metric. TABLE II shows the results of experiment 1 (using only the ontological similarity) while applying different window sizes, from 1 to 4 sentences per text block. From the results we can see that the error rates are not high for all the subsets (range from 0.15 to 0.32), which means that generating text segments based on the ontological relation between its constituents is feasible with low error rates. It can also be noticed that varying the window size does not increase the quality of the segmentation; in contrast, it decreases the quality for some subsets. *Fig. 4* depicts the impact of the window size on the quality of the produced segments.

The lowest error rates can be seen in the 9-11 subset (0.15 for all windows), while the highest error rates can be seen in the 3-5 subset. Intuitively, this implies that as the length of the reference segments (i.e. the real segments from the original text) increases, the efficiency of text segmentation increases. This implication reinforces the feasibility of our approach. This is because, as mentioned before, long segments exhibit more interlinking conceptual relations than short segments.

TABLE III shows the results of experiments 2, 3, and 4 where we evaluated the hybrid approach that combines the ontological and lexical similarities using different weights.

TABLE II.  ONTOLOGICAL SIMILARITY ERROR RATES (WD) FOR DIFFERENT WINDOW SIZES

| Range of *n* / Window | 3-11 | 3-5 | 6-8 | 9-11 |
|---|---|---|---|---|
| **W = 1** | 0.21 | 0.32 | 0.20 | 0.15 |
| **W = 2** | 0.21 | 0.32 | 0.21 | 0.15 |
| **W = 3** | 0.21 | 0.34 | 0.21 | 0.15 |
| **W = 4** | 0.22 | 0.34 | 0.21 | 0.15 |

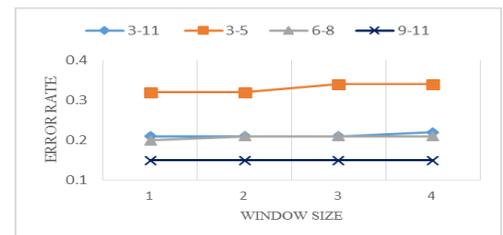

Fig. 4. Error rates of the Ontological Similarity using different window sizes

The results of experiment 2 indicate that when α = 0.3, the error rates of the segmentation in all the subsets are less than the error rates using the ontological similarity only (TABLE II). In experiments 3 and 4, we notice that as α increases (0.5 and 0.7 respectively), the error rates decrease. According to (5), when α increases, the lexical similarity weight is more than the ontological similarity weight, which means that combining the lexical similarity with the ontological similarity enhances the quality of the produced segments.

Furthermore, it is noticed that, as in experiment 1, when the window size increases, the error rate also increases which means that the segmentation quality decreases. The chart in *Fig. 5* illustrates that increasing the window size increases the error rate with α = 0.3. Charts for setting α = 0.5 and 0.7 are in APPENDIX I. *Fig. 6* depicts the error rates for the four experiments using window = 1 (charts for using other windows are in APPENDIX I).

To the best of our knowledge, there is no segmentation approach that uses ontological similarity to segment text. Therefore, it is not possible to compare the evaluation scores of our approach to those of a similar approach. Nevertheless we can compare it with state-of-the-art approaches based on the segmentation quality in general.

TABLE III. HYBRID APPROACH ERROR RATES FOR DIFFERENT WINDOW SIZES

| Range of n / Window | 3-11 | 3-5 | 6-8 | 9-11 |
|---|---|---|---|---|
| Experiment 2: α = 0.3 | | | | |
| W = 1 | 0.17 | 0.22 | 0.17 | 0.13 |
| W = 2 | 0.19 | 0.29 | 0.18 | 0.14 |
| W = 3 | 0.19 | 0.34 | 0.20 | 0.14 |
| W = 4 | 0.20 | 0.33 | 0.20 | 0.15 |
| Experiment 3: α = 0.5 | | | | |
| W = 1 | 0.16 | 0.21 | 0.16 | 0.12 |
| W = 2 | 0.18 | 0.27 | 0.17 | 0.12 |
| W = 3 | 0.19 | 0.33 | 0.19 | 0.13 |
| W = 4 | 0.20 | 0.33 | 0.19 | 0.14 |
| Experiment 4: α = 0.7 | | | | |
| **W = 1** | **0.15** | **0.19** | **0.15** | **0.11** |
| W = 2 | 0.17 | 0.25 | 0.16 | 0.12 |
| W = 3 | 0.18 | 0.33 | 0.19 | 0.13 |
| W = 4 | 0.20 | 0.33 | 0.20 | 0.14 |

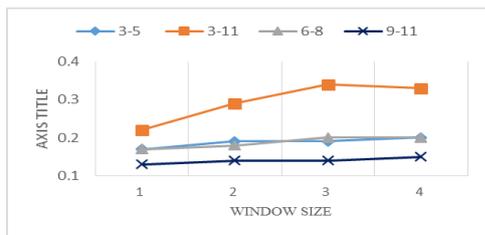

Fig. 5. The error rate of the Ontological and Lexical similarities for different window sizes with α = 0.3

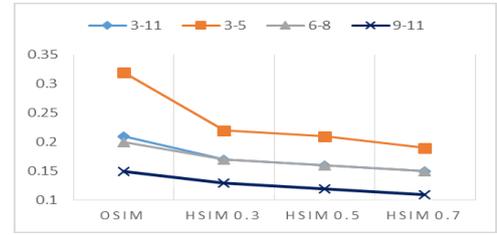

Fig. 6. The error rates for the four experiments with window = 1. (Hsim) is the Hybrid similarity (Ontological+ Lexical).

TABLE IV. $P_k$ VALUES FOR THE CHOI DATA SET FOR VARIOUS ALGORITHMS IN THE LITERATURE WITH PROVIDED SEGMENT NUMBER

| Approach | 3-11 | 3-5 | 6-8 | 9-11 |
|---|---|---|---|---|
| U00 | 0.11 | 0.13 | 0.06 | 0.06 |
| C99 | 0.13 | 0.18 | 0.10 | 0.10 |
| OntoSeg | 0.30 | 0.19 | 0.30 | 0.30 |
| TextTiling | 0.46 | 0.44 | 0.43 | 0.48 |

As we evaluated the performance of our approach using *WindowDiff*, we also evaluated it with *Pk*. The approaches that we compare our approach with were evaluated also with the *Pk* metric. Furthermore, these approaches were evaluated against the same dataset that we use in our experiments (Choi's dataset). Examples of such approaches are: *TextTiling* [8], *C99* [20], and *U00* [11]. TABLE IV presents a comparison of the performance of our approach compared to these approaches where number of segments needed is provided.[5]

Although *OntoSeg* (i.e. our segmentation approach that is based on ontological similarity) does not produce the best scores, the results show that it –as a novel method in text segmentation– is generally performing as good as current state-of-the-art approaches. In other words, the experimental results show that using ontological similarity in text segmentation is very promising and also that text segmentation can be performed in a way that does not depend on text (lexical) characteristics. This renders *OntoSeg* an approach that lends itself well to Ontology-based NLP tasks.

## VII. CONCLUSION AND FUTURE WORK

Text Segmentation (TS) is an essential pre-step for many Natural Language Processing (NLP) tasks, such as Information Retrieval and Text Summarisation. As these tasks are moving towards the Semantic Web and the use of Ontologies (e.g. ontology-based IR systems), this generates a need to segment text in a way that suits these ontology-based tasks. In this paper we presented a new approach to text segmentation based on the ontological similarity between text blocks. The proposed approach uses a Hierarchical Agglomerative Clustering (HAC) approach to iteratively cluster text segments that are deemed to be ontologically similar to each other. The output is a tree-like hierarchy of the text. We showed that the produced hierarchy is beneficial in producing hierarchical text segments with different levels of granularity, and also in producing linear text segments by flattening the obtained tree. The results of our experiments showed that using ontological similarity (even on its own) performs successful segmentation with low error rates; this reflects that the ontological segmentation approach has good potential for being used in modern ontology-based systems. We also enhanced the proposed approach by combining the lexical similarity with the ontological similarity; to this end, the experimental results showed that this combination enhanced the produced segments.

---
[5] The results were brought from Utiyama and Isahara [11] & Riedl and Biemann [39] papers.

Moving forward, viable future work may involve examining a number of factors that can enhance the segmentation process. For example, it is expected that the choice of the knowledge base ontology to use definitely affects the segmentation quality; the richness of the ontology reflects the richness of the semantic annotation of text. Also the ontology-based similarity approach represents an important factor in enhancing the segmentation quality. As mentioned before, there are different approaches to measure the similarity between two concepts in an ontology, of which we used the edge-counting based approach. For the other approaches that rely on concept properties and Information Content (IC), they measure the similarity between concepts from different perspectives, and provide, for a concept, a better understanding of its semantics. Using these approaches in the similarity measurement may contribute to improving the segmentation quality.

ACKNOWLEDGMENT

This work is supported by Science Foundation Ireland (Grant 12/CE/I2267) as part of CNGL Centre for Global Intelligent Content (www.cngl.ie) at Trinity College Dublin.

## APPENDIX I

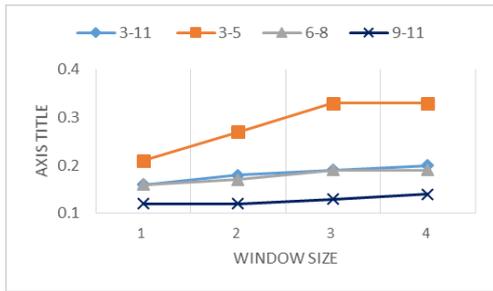

Fig. 7. The error rate of the Ontological and Lexical similarities for different window sizes with α = 0.5

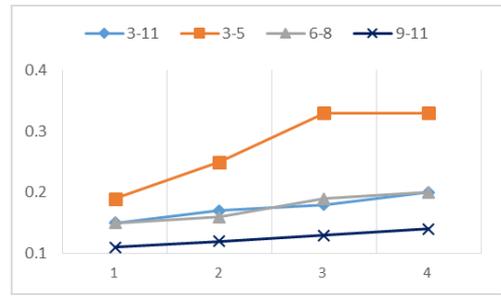

Fig. 8. The error rate of the Ontological and Lexical similarities for different window sizes with α = 0.7

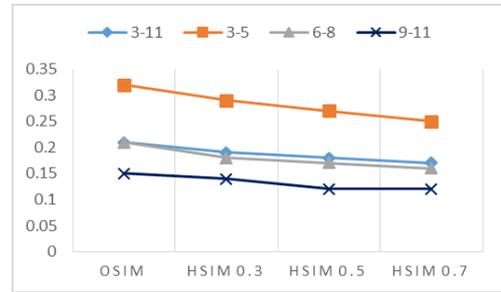

Fig. 9. The error rates for the four experiments with window = 2.

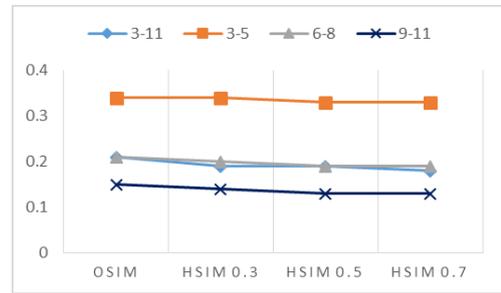

Fig. 10. The error rates for the four experiments with window = 3.

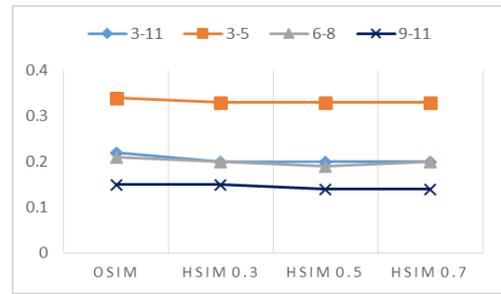

Fig. 11. The error rates for the four experiments with window = 4.